\documentclass[runningheads]{llncs}

\usepackage{eccvabbrv}

\usepackage{graphicx}
\usepackage{booktabs}

\usepackage[accsupp]{axessibility}  

\usepackage{wrapfig}
\usepackage{amsmath}
\usepackage{amssymb}
\usepackage{mathtools}
\usepackage{multirow}
\usepackage{pifont}
\usepackage{array}     
\usepackage{dsfont}

\usepackage{microtype}
\usepackage[table]{xcolor} 
\definecolor{grayhighlight}{gray}{0.9}

\definecolor{dbcolor}{rgb}{0.5,0,1}

\usepackage{hyperref}

\begin{document}

\title{Enhancing Alignment for Unified Multimodal Models via Semantically-Grounded Supervision} 

\titlerunning{Semantically-Grounded Supervision (SeGroS)}

\author{Jiyeong Kim\textsuperscript{1}\and
Yerim So\textsuperscript{1}\and
Hyesong Choi\textsuperscript{2}\and
Uiwon Hwang\textsuperscript{1}\and
Dongbo Min\textsuperscript{1,$\dagger$}}
\authorrunning{J.~Kim et al.}

\institute{\textsuperscript{1} Ewha Womans University, South Korea \\ \textsuperscript{2} Soongsil University, South Korea \\
\email{\{wldud8946, yrso, uiwon.hwang, dbmin\}@ewha.ac.kr},  \email{hyesong@ssu.ac.kr} \\}

\maketitle
\begingroup
\renewcommand{\thefootnote}{}
\footnotetext{\textsuperscript{$\dagger$} Corresponding author.}
\endgroup
\begin{abstract}
  Unified Multimodal Models (UMMs) have emerged as a promising paradigm that integrates multimodal understanding and generation within a unified modeling framework. However, current generative training paradigms suffer from inherent limitations. We present {\bf Se}mantically-{\bf Gro}unded {\bf S}upervision ({\bf SeGroS}), a fine-tuning framework designed to resolve the granularity mismatch and supervisory redundancy in UMMs. At its core, we propose a novel visual grounding map to construct two complementary supervision signals. First, we formulate semantic \textit{Visual Hints} to compensate for the sparsity of text prompts. Second, we generate a semantically-grounded \textit{Corrupted Input} to explicitly enhance the supervision of masking-based UMMs by restricting the reconstruction loss to core text-aligned regions. Extensive evaluations on GenEval, DPGBench, and CompBench demonstrate that SeGroS significantly improves generation fidelity and cross-modal alignment across various UMM architectures.
\end{abstract}
    
\section{Introduction}

The evolution of Multimodal Large Language Models (MLLM)~\cite{liu2023visual,bai2023qwen} has shifted the prevailing paradigm from text-only processing~\cite{devlin2019bert,brown2020language} to multimodal reasoning, enabling models to jointly interpret visual and linguistic signals. 
In extending to handle both understanding and image generation, early efforts often relied on combining independent models (\eg, an LMM for reasoning and a dedicated image generator)~\cite{wu2024next,koh2023generating}.
However, such decoupled designs suffer from weak integration between modalities~\cite{wu2025harmonizing} and incur engineering costs to maintain two separate models independently~\cite{wu2024vila}.
This led to the emergence of {\bf U}nified {\bf M}ultimodal {\bf M}odels ({\bf UMM}s)~\cite{team2024chameleon,ge2024seed,liao2025mogao,liu2025tuna,sun2023emu,lin2025bifrost}, which integrate multimodal understanding and generation within a single sequence modeling framework by representing visual signals as tokenized inputs analogous to text tokens.

\begin{figure*}[t] 
\begin{center}
\includegraphics[width=1.0\linewidth]{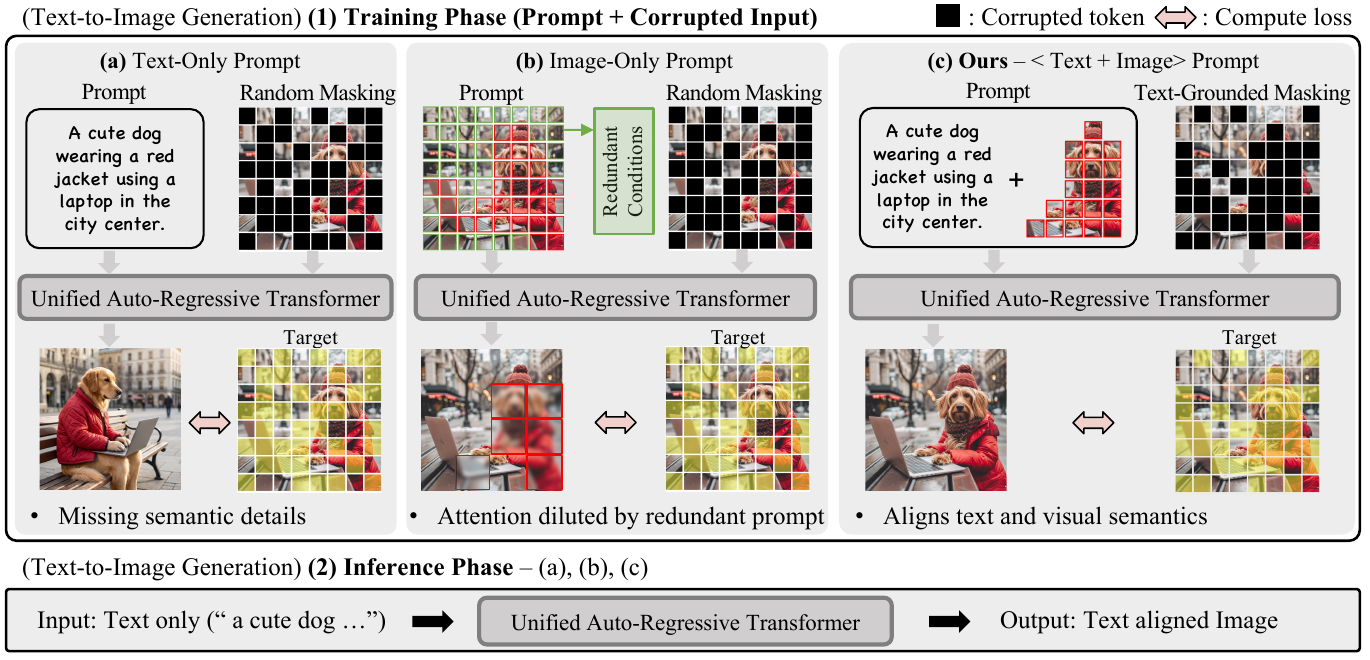}
\end{center}
\caption{\textbf{Generative Training Paradigms in UMMs.}
(a) Text-conditioned training in UMMs~\cite{wu2025harmonizing,xieshow} uses the text prompt as the semantic specification and applies random masking to form the corrupted input for token-level reconstruction. 
(b) Image reconstruction-based approach, Reca~\cite{xie2025reconstruction}, uses image prompts to reconstruct corrupted input, yet still relies on random masking.
(c) Our approach, {\bf Se}mantically-{\bf Gro}unded {\bf S}upervision ({\bf SeGroS}), constructs its training signal by conditioning on text and explicitly text-aligned visual prompts. Furthermore, we selectively mask the core semantic regions to construct the corrupted input. This provides highly efficient supervision that concentrates model capacity on core structures rather than irrelevant regions.
}
\label{fig:intro}
\end{figure*}

Early UMMs~\cite{team2024chameleon,ge2023making,chen2025janus} primarily adopt standard next-token prediction, generating visual tokens autoregressively in a unidirectional sequence. However, this sequential nature is computationally inefficient~\cite{chang2022maskgit,chang2023muse,lezama2022improved} and struggles to capture complex 2D spatial dependencies required for holistic 2D visual modeling~\cite{tian2024visual,li2024autoregressive,xieshow}. To overcome these limitations, subsequent approaches shift toward mask prediction or denoising objectives, reconstructing corrupted visual tokens conditioned on both text and visible visual context.
For instance, Show-o~\cite{xieshow} introduces an \textbf{AR-diffusion} architecture that integrates autoregressive (AR) modeling with masked diffusion over visual tokens, enabling parallel decoding and faster generation.  Harmon~\cite{wu2025harmonizing} incorporates masked autoregressive (MAR) modeling~\cite{li2024autoregressive} within an \textbf{AR-MAR} framework to reconstruct corrupted visual tokens. 
Despite architectural differences, these paradigms share a common training strategy: the model processes a \textit{corrupted input} in which part of the visual tokens are masked and the rest remain visible, and computes the reconstruction loss only on the masked subset (Fig.~\ref{fig:intro} (a)). Reconstructing these tokens conditioned on the \textit{text prompt} provides a cross-modal learning signal.

Despite architectural progress, a fundamental bottleneck still persists across UMM paradigms: the granularity mismatch between textual conditions and visual targets. While visual tokens encode dense spatial structure and fine-grained details, text prompts provide only abstract semantic constraints~\cite{hahn2025proactive,xie2025reconstruction,lin2025bifrost}. As a result, a single text description can correspond to multiple visually distinct images, rendering text-to-image supervision inherently ambiguous. As illustrated in Fig.~\ref{fig:intro} (a), the text prompt typically specifies primary objects (\eg, a dog wearing a red jacket) but leaves specific attributes (\eg, texture, illumination, pose, and spatial layout) underspecified. During training, however, the model is required to reconstruct one specific target image. Consequently, even semantically valid visual variations may be penalized if they do not exactly match the \textbf{text-underspecified factors} of the target, encouraging the model to fit incidental instance-level details rather than learning robust semantic alignment.

To compensate for missing fine-grained visual details in text-only conditioning, recent UMM training paradigms incorporate reconstruction objectives based on image-conditioned prompts (\ie, \textit{Visual Hints})~\cite{xie2025reconstruction}, as illustrated in Fig.~\ref{fig:intro} (b). By leveraging visual tokens from the image, this method provides dense conditioning cues that are absent in the text prompts. 
However, in image-conditioned training, using all visual tokens as visual hints can be redundant, as many patches correspond to low-salience background regions that dilute attention and weaken semantic alignment.
Furthermore, regardless of whether conditioning is provided by text~\cite{xieshow,wu2025harmonizing} or image prompts~\cite{xie2025reconstruction}, existing UMMs typically define the corrupted input via random masking that is agnostic to semantic salience.
Consequently, a substantial portion of reconstruction supervision is wasted on semantically irrelevant regions, encouraging the model to fit incidental visual details rather than strengthening text-aligned visual concepts.

In this work, we introduce {\bf Se}mantically-{\bf Gro}unded {\bf S}upervision ({\bf SeGroS}), a fine-tuning framework for UMMs that mitigates inefficient supervision arising from the granularity mismatch between coarse text and dense visual tokens.
As illustrated in Fig.~\ref{fig:intro} (c), SeGroS departs from previously used conditioning training by providing semantically grounded visual guidance during training.
At the core of SeGroS is a visual grounding score that explicitly quantifies the alignment between discriminative text tokens and image patches, allowing us to construct structured supervision.
Based on this score, we derive two complementary components: \textit{Visual Hints}, which provide semantically grounded visual embeddings to guide generation, and semantically grounded \textit{Corrupted Input}, which allocates reconstruction supervision to text-aligned regions, reducing redundant loss on irrelevant background details (Fig.~\ref{fig:method}).

The proposed framework consists of three key steps. We first perform \textbf{Discriminative Text Token Filtering} to identify linguistically salient tokens that have strong visual counterparts. 
Since uniformly comparing all tokens dilutes the alignment scores of core regions, we compute both intra-modal (text–text) and inter-modal (text–image) affinities to retain text tokens that dominate the visual semantics.
Then, we construct a \textbf{Visual Grounding Map} by measuring the similarity between the \emph{filtered} text tokens and each image patch. This map captures fine-grained text–image correspondences and identifies semantically coherent visual regions. 
Finally, guided by this map, patches with high grounding scores are explicitly extracted to serve as \textit{Visual Hints}. Meanwhile, patches with the lowest scores are selected to form the unmasked context of the semantically-grounded \textit{Corrupted Input}. This naturally leaves the core semantic areas masked, forcing the model to actively reconstruct them rather than random patches.

Our main contributions are fourfold:
{\bf (1)} We propose SeGroS, a semantically grounded fine-tuning framework for UMMs that enhances cross-modal alignment by overcoming the text-image granularity mismatch. {\bf (2)} Within SeGroS, we introduce a fine-grained grounding mechanism that first filters discriminative text tokens and then constructs a visual grounding map from these filtered tokens to extract text-aligned image regions. {\bf (3)} Leveraging this grounding, we construct \textit{Visual Hints} for conditioning and a semantically grounded \textit{corrupted input}, concentrating supervision on core semantic regions.
{\bf (4)} We demonstrate that SeGroS improves generation fidelity and cross-modal alignment across diverse UMMs, achieving strong results on GenEval, DPGBench, and CompBench.

\section{Preliminary}
We formulate unified multimodal generation and understanding under masked/ denoising UMM paradigms~\cite{xieshow,wu2025harmonizing}.
To maintain an architecture-agnostic formulation, we abstract away model-specific instantiations (e.g., masked diffusion denoising~\cite{xieshow} or masked autoregressive decoding~\cite{wu2025harmonizing}) and instead focus on their shared training principle: reconstructing masked visual tokens from the corrupted input conditioned on the text prompt.

\noindent{\bf {Unified Input Representations.}}
We define all input modalities in a shared $D$-dimensional latent space.
\textbf{Text.} Given a text prompt $\mathbf{T}$, we tokenize it into text tokens and map them to embeddings via a pre-trained embedding table, yielding $\mathbf{Z}_T\in\mathbb{R}^{L_T\times D}$, where $L_T$ denotes the text sequence length. 
\textbf{Vision.} 
Given an image $\mathbf{I}$, a pretrained tokenizer/encoder yields either discrete tokens or continuous latents, which we represent in the shared space as $\mathbf{Z}_I \in \mathbb{R}^{N_I \times D}$, where $N_I$ denotes the number of visual tokens.

\noindent{\bf Corrupted Input.}
Following standard practices, at each training step we randomly choose a masking ratio
$\gamma(t) \in [0.7, 1.0)$, where $\gamma(t)$ denotes the fraction of tokens treated as unknown
(\eg, $\gamma(t)=0.7$ masks 70\% of tokens).
We then sample a mask index set $\mathcal{M} \subset \{1,\dots,N_I\}$ with
$|\mathcal{M}| = \lfloor \gamma(t)\, N_I \rfloor$.
The masked and unmasked subsets are defined as
$\mathbf{Z}_I^{\text{mask}} = \{ z_i \mid i \in \mathcal{M} \}$ and
$\mathbf{Z}_I^{\text{seen}} = \{ z_i \mid i \notin \mathcal{M} \}$.
The corrupted input $\widetilde{\mathbf{Z}}_I$ is constructed by replacing tokens in
$\mathbf{Z}_I^{\text{mask}}$ with a learnable \texttt{[MASK]} embedding, while keeping
$\mathbf{Z}_I^{\text{seen}}$ unchanged.
This formulation reflects the common mask-and-reconstruct training paradigm in masked/denoising UMMs,
with supervision restricted to the masked subset.

\noindent{\bf {Training Objective.}}
The unified model $f_\theta$ is optimized for Text-to-Image (T2I) generation by minimizing a generalized reconstruction loss:
\begin{equation}
\mathcal{L}_{\text{t2i}} = \ell\big(f_\theta(\mathbf{Z}_T, \widetilde{\mathbf{Z}}_I), \, \mathbf{Z}_{I}\big),
\label{eq:t2i_base}
\end{equation}
where the loss $\ell(\cdot)$ is computed only over the masked subset $\mathcal{M}$.
The loss $\ell(\cdot,\cdot)$ is instantiated as negative log likelihood for discrete tokens (e.g., Show-o~\cite{xieshow}) or Mean Squared Error (MSE) for continuous latents (e.g., Harmon~\cite{wu2025harmonizing}).

Furthermore, to enable multimodal understanding and text generation capabilities, we incorporate an autoregressive Image-to-Text (I2T) objective:
\begin{equation}
\mathcal{L}_{\text{i2t}}
= \ell\big(f_\theta(\mathbf{Z}_I, \mathbf{Z}_{T,[\texttt{question}]}), \mathbf{Z}_{T,[\texttt{answer}]}\big),
\end{equation}
where $\ell$ is the standard negative log-likelihood loss for autoregressively predicting the answer $\mathbf{Z}_{T,[\texttt{answer}]}$ conditioned on the image $\mathbf{Z}_I$ and question $\mathbf{Z}_{T,[\texttt{question}]}$.
\section{Proposed Method}
\subsection{Motivation and Overview}

\begin{figure*}[t] 
\begin{center}
\includegraphics[width=1.0\linewidth]{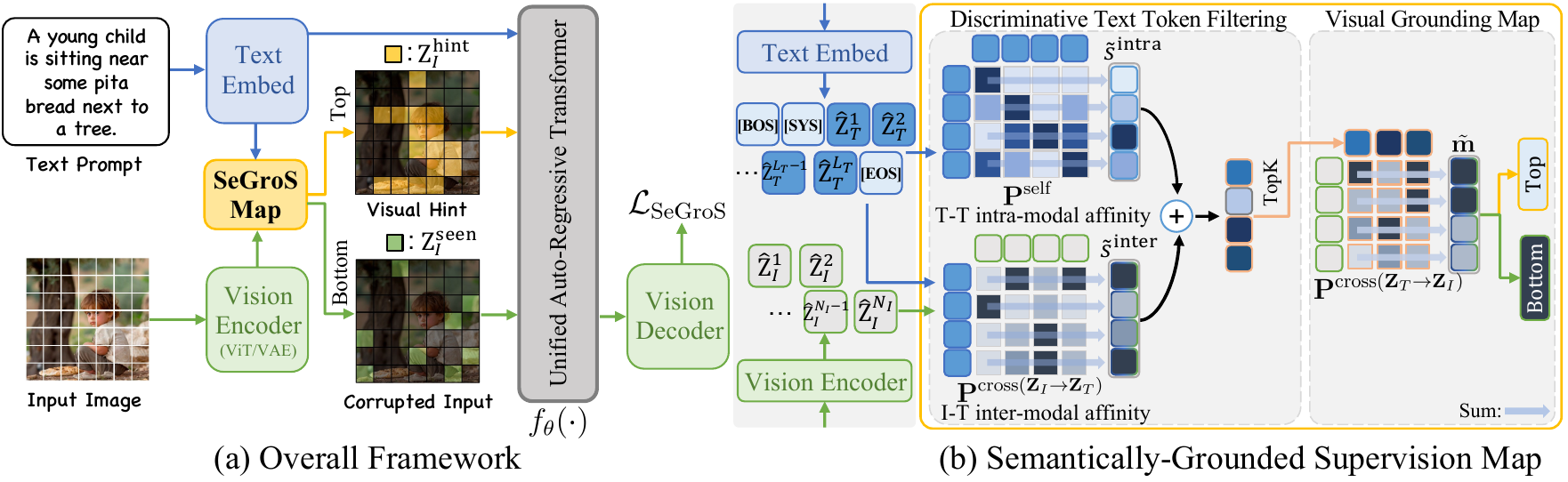}
\end{center}
\caption{{\bf Overview of SeGroS during training.} 
(a) The SeGroS Map computes a grounding score for each visual token based on its semantic similarity to discriminative text tokens and constructs two complementary training signals. The \textit{Visual Hints} preserve high-groundedness tokens as additional conditioning signals (yellow patches). The \textit{Corrupted Input} retains low-groundedness tokens as unmasked context (green patches) while masking the remaining tokens, which are reconstructed by the unified autoregressive Transformer. (b) Discriminative Text Token Filtering identifies visually relevant text tokens via intra- and inter-modal affinities, which are then used to compute the Visual Grounding Map. At inference time, SeGroS Map generation is omitted, allowing the original UMMs to generate images without architectural modifications.
}
\label{fig:method}
\end{figure*}
Under text-only conditioning, Eq.~\eqref{eq:t2i_base} suffers from a \textit{granularity mismatch}: sparse text lacks fine-grained constraints~\cite{hahn2025proactive,xie2025reconstruction,lin2025bifrost}, resulting in ambiguous reconstruction supervision (Fig.~\ref{fig:intro} (a)).
To address the sparsity issue of text prompts, Reca~\cite{xie2025reconstruction} introduces image-only conditioning, leveraging dense visual tokens as prompts, as described in Fig.~\ref{fig:intro} (b):
$\mathcal{L}_{\text{reca}} = \ell\big(f_\theta(\mathbf{Z}_I, \widetilde{\mathbf{Z}}_I), \, \mathbf{Z}_{I}\big)$.
The corrupted input $\widetilde{\mathbf{Z}}_I$ is reconstructed conditioned on the visual tokens $\mathbf{Z}_I$ (visual hint).
However, since the image prompt $\mathbf{Z}_I$ also contains irrelevant background regions, conditioning on all visual tokens introduces \textit{supervisory redundancy}, which can dilute attention and weaken semantic alignment. Furthermore, semantic-agnostic random masking often assigns reconstruction targets to low-salience background regions, wasting supervision on incidental details rather than strengthening text-aligned structure.

To empirically validate this hypothesis, we conducted an ablation on the image-only prompting setting of Reca~\cite{xie2025reconstruction} to test whether conditioning on all visual tokens is necessary.
We ranked image patches by a text–image affinity score (used only for patch selection) and kept only the top 30\% as visual hints. Despite using 3.3× fewer hint tokens, this prompt pruning improves GenEval~\cite{ghosh2023geneval} from 78.7→79.2 and DPGBench~\cite{hu2024ella} from 84.7→85.2 on the Harmon 0.5B model~\cite{wu2025harmonizing}.
These preliminary results explicitly validate our claim: dense image prompts inherently suffer from background redundancy, proving that semantically selective conditioning on text-grounded regions is essential.

Driven by this finding, we propose {\bf Se}mantically-{\bf Gro}unded {\bf S}upervision ({\bf SeGroS}), a fine-tuning framework that restructures dense visual supervision according to text–image semantic correspondence (Fig.~\ref{fig:method}). We explicitly identify which image regions are semantically grounded in the text and allocate supervision accordingly. SeGroS operates in three conceptual steps. First, we perform \textbf{Discriminative Text Token Filtering} to extract linguistically salient tokens that are likely to have visual counterparts, retaining key entities and interactions that dominate the scene semantics. Second, we construct a \textbf{Visual Grounding Map} that measures patch-level similarity between the filtered text tokens and visual features, quantifying how strongly each image patch aligns with the textual semantics. Finally, based on the grounding map, we construct two complementary supervision signals: (1) \textit{Semantic Visual Hints}, consisting of highly aligned regions provided as explicit conditioning signals, and (2) \textit{Semantically grounded Corrupted Input}, which is input for reconstruction. 
Specifically, low-groundedness tokens are retained as unmasked context, while high-groundedness tokens are masked and reconstructed by the unified autoregressive Transformer. 
By concentrating the reconstruction objective on semantically coherent regions, SeGroS provides structured supervision that strengthens cross-modal alignment while reducing redundant visual noise.

\subsection{Discriminative Text Token Filtering}
We first refine the text prompt $\mathbf{Z}_T$ by identifying tokens that are most relevant for visual grounding, as treating all text tokens equally can dilute alignment scores for core visual regions. A text token is considered discriminative only if it is both linguistically salient and visually grounded. To this end, we evaluate each token from two complementary perspectives. We compute the \textbf{Text Intra-modal Affinity} to measure its global linguistic importance within the text prompt, and the \textbf{Text–Image Inter-modal Affinity} to assess its correspondence with visual tokens. Tokens that score highly on both criteria are retained as discriminative text tokens for subsequent groundedness estimation.

\subsubsection{Text Intra-modal Affinity.}
The first step is to identify the core concepts within the text prompt itself.
Given the $\ell_2$-normalized text embeddings $\hat{\mathbf{Z}}_T \in \mathbb{R}^{L_T \times D}$ extracted by the tokenizer prior to processing by the unified multimodal model $f_\theta$,
we compute the self-affinity matrix $\mathbf{S} \in \mathbb{R}^{L_T \times L_T}$ as
$\mathbf{S}=\hat{\mathbf{Z}}_T{\hat{\mathbf{Z}}_T}^\top$ to capture contextual
relationships between tokens. 
To focus on core semantic concepts, we exclude special tokens (\eg, \texttt{[BOS]}, \texttt{[EOS]}) from the calculation, which are known to act as attention sinks~\cite{xiao2023efficient,kim2025text} that bias attention weights away from discriminative content.
We then derive the attention probability map $\mathbf{P}^{\text{self}}$ by applying a row-wise softmax with a temperature parameter $\tau$:
\begin{equation}
    \mathbf{P}^{\text{self}}_{kj} = \frac{\exp(\mathbf{S}_{kj} / \tau)}{\sum_{m=1}^{L_T} \exp(\mathbf{S}_{km} / \tau)}.
\end{equation}
Finally, the intra-modal affinity score $s^{\text{intra}}_j$ for the $j$-th text token is obtained by aggregating the attention it receives from all other tokens in the sequence:
\begin{equation}
s^{\text{intra}}_j = \sum_{k=1}^{L_T} \mathbf{P}^{\text{self}}_{kj}.
\end{equation}

\subsubsection{Image-Text Inter-modal Affinity.}
While intra-modal affinity identifies semantically dominant tokens, it does not guarantee their visual congruence with the actual image content.
To address this, we evaluate the visual relevance of each text token by measuring its direct influence on the image features.
We compute the image-text inter-modal affinity matrix $\mathbf{A} \in \mathbb{R}^{N_I \times L_T}$ as $\mathbf{A}=\hat{\mathbf{Z}}_I\hat{\mathbf{Z}}_T^\top$, using the $\ell_2$-normalized visual token embeddings $\hat{\mathbf{Z}}_I$ and text token embeddings $\hat{\mathbf{Z}}_T$, where both are derived from their respective tokenizers prior to processing by the unified multimodal model $f_\theta$.
Similar to the text intra-modal affinity, we calculate the inter-modal attention probability $\mathbf{P}^{\text{cross}(\mathbf{Z}_I \to \mathbf{Z}_T)}$ that represents the likelihood of visual token $i$ attending to text token $j$:
\begin{equation}
    \mathbf{P}^{\text{cross}(\mathbf{Z}_I \to \mathbf{Z}_T)}_{ij} = \frac{\exp(\mathbf{A}_{ij} / \tau)}{\sum_{k=1}^{L_T} \exp(\mathbf{A}_{ik} / \tau)}.
\end{equation}
The inter-modal affinity score $s^{\text{inter}}_j$ is defined by aggregating the attention the $j$-th text token receives from the entire visual context. This metric filters out text concepts that have little visual grounding in the image:
\begin{equation}
    s^{\text{inter}(\mathbf{Z}_I \to \mathbf{Z}_T)}_j = \sum_{i=1}^{N_I} \mathbf{P}^{\text{cross}}_{ij}.
\end{equation}
After computing the intra-modal centrality ($s^{\text{intra}}$) and inter-modal relevance ($s^{\text{inter}(\mathbf{Z}_I \to \mathbf{Z}_T)}$), we integrate these metrics to determine the final semantic importance of each text token.

\subsubsection{Discriminative Token Selection.} Finally, to identify the most discriminative tokens, we integrate the linguistic and visual insights derived above. Since the two affinity scores operate on different scales, we normalize them via Min-Max scaling to ensure a balanced contribution.
For each score vector
$\mathbf{s} \in \{\mathbf{s}^{\text{intra}}, \mathbf{s}^{\text{inter}(\mathbf{Z}_I \to \mathbf{Z}_T)}\}$, we compute
$\tilde{s}_j = \frac{s_j - \min(\mathbf{s})}{\max(\mathbf{s}) - \min(\mathbf{s})}$.
The unified importance score $\Omega_j$ is then defined as the sum of the normalized metrics:
\begin{equation}
\Omega_j = \tilde{s}_j^{\text{intra}} + \tilde{s}_j^{\text{inter}}.
\end{equation}

We select the indices $\mathcal{K}$ of the top $K_T$ tokens, where $K_T = \lfloor \rho L_T \rfloor$ is determined by the sequence length $L_T$ and a text preservation ratio $\rho$:
\begin{equation}
\mathcal{K} = \text{TopK}(\Omega, K_T) = \{j \mid \text{rank}(\Omega_j) \le K_T \}
\label{eq:topk_text}
\end{equation}
Accordingly, we define the binary text-filtering mask $\mathbf{w} \in \{0, 1\}^{L_T}$:
\begin{equation}
    w_j = \mathds{1}(j \in \mathcal{K}) = \begin{cases} 1 & \text{if } j \in \mathcal{K} \\ 0 & \text{otherwise}. \end{cases}
\end{equation}

\subsection{Visual Grounding Map}
With the binary text-filtering mask $\mathbf{w}$ identified, we evaluate how strongly each visual token aligns with the filtered text tokens, yielding the Visual Grounding Map. This map serves as the decisive criterion for constructing two complementary training signals:
\textit{Visual Hints} and \textit{Corrupted Input}.
Similar to the image-to-text affinity, we calculate the text-to-image attention probability $\mathbf{P}^{\text{cross}(\mathbf{Z}_T \to \mathbf{Z}_I)}$ that represents the likelihood of text token $j$ attending to visual token $i$:
\begin{equation}
\mathbf{P}^{\text{cross}(\mathbf{Z}_T \to \mathbf{Z}_I)}_{ji} = \frac{\exp(\mathbf{A}^\top_{ji} / \tau)}{\sum_{k=1}^{N_I} \exp(\mathbf{A}^\top_{jk} / \tau)}.
\end{equation}
Using this probability, we compute the visual grounding score $m_i$ for each image patch $i$ by aggregating the text-to-image attention probabilities weighted by the text filtering mask $\mathbf{w}$:
\begin{equation}
    m_i = \sum_{j=1}^{L_T} w_j \cdot \mathbf{P}^{\text{cross}(\mathbf{Z}_T \to \mathbf{Z}_I)}_{ji},
    \quad \text{for } i \in \{1,\dots,N_I\}.
\end{equation}

These scores are collected into a vector $\mathbf{m}=[m_1,\dots,m_{N_I}] \in \mathbb{R}^{N_I}$, which is then normalized to the $[0, 1]$ range via Min--Max scaling, defined as $\bar{m}_i = \frac{m_i - \min(\mathbf{m})}{\max(\mathbf{m}) - \min(\mathbf{m})}$.
Directly applying deterministic selection to this score map enforces a fixed construction of \textit{Visual Hints} and \textit{Corrupted Input}, repeatedly isolating the exact same regions across epochs. While text strictly requires such determinism to preserve core semantics, images are fundamentally spatially redundant, often containing multiple patches with similar grounding scores. Thus, exclusive reliance on deterministic masking leads to a performance decline~\cite{choi2024salience}. To mitigate this, we inject a slight uniform noise $\boldsymbol{\xi} \sim \mathcal{U}([0, 0.5]^{N_I})$ into the normalized visual scores $\bar{\mathbf{m}}$ before the Top/Bottom selection, formulated as:
\begin{equation}
    \tilde{\mathbf{m}} = \bar{\mathbf{m}} + \boldsymbol{\xi}.
\label{eq:noise_score}
\end{equation}

\subsection{Visual Hints and Corrupted Input Construction}
Based on the perturbed visual grounding map $\tilde{\mathbf{m}}$, we map the visual tokens to their respective roles using two distinct criteria, as shown in Fig.~\ref{fig:method}.
First, to resolve the linguistic ambiguity of the sparse text prompt, we construct \textit{Visual Hints} ($\mathbf{Z}_I^{\text{hint}}$) to serve as an explicit visual prompt. Specifically, given a predefined selection ratio $\eta$, we define these hints by selecting the top-$\lfloor \eta N_I \rfloor$ tokens with the highest scores, representing the visual parts most highly aligned with the text. Our empirical analysis (Tab.~\ref{tab:ablation_ratio_simple}) identifies the range $\eta \in [0.3, 0.5]$ as the optimal balance for providing effective semantic cues; thus, we adopt $\eta\in$[0.3, 0.4].

Second, we define a semantically grounded \textit{Corrupted Input}
($\widetilde{\mathbf{Z}}_I^{\text{SeGroS}}$) to focus reconstruction on text-aligned regions.
Instead of randomly sampling a mask set~\cite{xieshow,xie2025reconstruction,wu2025harmonizing}, we deliberately retain semantically low-salience regions as visible context and mask the core semantic regions as reconstruction targets.
Given a masking ratio $\gamma(t)\in[0.7,1.0)$, we set
$K_{\text{mask}}=\lfloor \gamma(t) N_I \rfloor$ and $K_{\text{seen}}=N_I-K_{\text{mask}}$.
We select the bottom-$K_{\text{seen}}$ indices from the grounding map $\tilde{\mathbf{m}}$ to remain visible:
$\mathcal{S}=\text{BottomK}(\tilde{\mathbf{m}}, K_{\text{seen}})$,
and define $\mathbf{Z}_I^{\text{seen}}=\{z_i\mid i\in\mathcal{S}\}$.
The mask index set is then given by $\mathcal{M}=\{1,\dots,N_I\}\setminus\mathcal{S}$,
and the corrupted input is formed by replacing $\{z_i\mid i\in\mathcal{M}\}$ with a learnable \texttt{[MASK]} embedding.

\subsubsection{Final Objective.}
We optimize a joint objective where \textit{Visual Hints} provide additional conditioning alongside the text prompt, and the semantically grounded \textit{Corrupted Input} specifies the masked reconstruction targets.
Conditioned on $\mathbf{Z}_T$ and $\mathbf{Z}_I^{\text{hint}}$, the model $f_\theta$ processes the partially masked sequence $\widetilde{\mathbf{Z}}_I^{\text{SeGroS}}$ and reconstructs the original visual tokens at the masked positions.
To preserve the model's image-to-text (I2T) capability, we additionally include an autoregressive I2T objective. The total loss is:
\begin{equation}
\mathcal{L}_{\text{total}}
=
\underbrace{\ell\big(f_\theta(\mathbf{Z}_T, \mathbf{Z}_I^{\text{hint}}, \widetilde{\mathbf{Z}}_I^{\text{SeGroS}}), \mathbf{Z}_I\big)}_{\mathcal{L}_{\text{SeGroS}}}
+ \lambda \mathcal{L}_{\text{i2t}},
\label{eq:final_objective}
\end{equation}
where $\ell(\cdot,\cdot)$ is evaluated only on the masked subset of $\widetilde{\mathbf{Z}}_I^{\text{SeGroS}}$. $\lambda$ is a fixed hyperparameter that balances the two loss terms.

\section{Experiments}
\subsection{Settings}
\subsubsection{Unified Multimodal Models (UMMs).}
To demonstrate the robustness of SeGroS, we conducted extensive experiments across three families of UMMs, encompassing diverse model scales (from 0.5B to 3.6B parameters) and generation resolutions (256$\times$256 and 512$\times$512). All models were initialized with their official pre-trained checkpoints before fine-tuning. The details are as follows: 

\noindent\textbf{(I) Show-o}~\cite{xieshow} unifies a masking-based diffusion model within a single autoregressive framework (\textbf{AR+Diffusion}). We adopted both Show-o-256 and Show-o-512 variants to evaluate performance across different image resolutions.
\noindent\textbf{(II) Harmon}~\cite{wu2025harmonizing} integrates a Masked Autoregressive (MAR)-based generative model into a unified autoregressive architecture (\textbf{AR+MAR}). We utilized Harmon-0.5B and Harmon-1.5B to assess the scalability of our method across varying model capacities.
\noindent\textbf{(III) OpenUni}~\cite{wu2025openuni} integrates an autoregressive understanding model with a diffusion-based generation model (\textbf{AR+Diffusion}). To validate our approach, we adopt the base OpenUni models (1.6B and 3.6B) operating at a $512\times512$ resolution, excluding the GPT-4o-Image distillation.
\begin{table*}[t]
    \centering
    \renewcommand{\arraystretch}{0.9} 
    \setlength{\tabcolsep}{3.5pt} 
    \caption{\textbf{Comparison on text-to-image generation across UMMs.} We compared the proposed method with the Supervised Fine-Tuning (SFT) and Reca~\cite{xie2025reconstruction}.
    `*' indicates results reproduced using the official implementations.}
    \label{tab:main_results}
    
    \small 
    \resizebox{1.0\linewidth}{!}{
    \begin{tabular}{l l ccccccc c c}
        \toprule
        \multirow{2}{*}{\textbf{UMMs}} & \multirow{2}{*}{\textbf{Method}} & \multicolumn{7}{c}{\textbf{GenEval} $\uparrow$} & \textbf{DPGBench} $\uparrow$ & \textbf{CompBench} $\uparrow$\\
        \cmidrule(lr){3-9} \cmidrule(lr){9-10} \cmidrule(lr){10-11}
         & & Single & Two & Count & Color & Position & Attr. & \textbf{Overall} & \textbf{Overall} & \textbf{Overall} \\
        \midrule
        \multicolumn{11}{l}{\cellcolor{gray!10}\textbf{\textit{Fine-Tuning Dataset: MidjourneyV6 Dataset}}} \\ 
        \midrule
        
        \multirow{4}{*}{Show-o-256~\cite{xieshow}} 
         & w/o SFT & 97.4 & 63.3 & 52.1 & 82.3 & 14.2 & 30.3 & 56.6 & 70.65 & 73.58\\
         & SFT*     & 97.8 & 63.4 & 53.4 & 80.3 & 16.5 & 33.5 & 57.5 & 74.6 & 75.37\\
         & Reca    & 97.4 & {\bf 73.6} & {\bf 56.0} & 83.8 & 20.3 & 40.2 & 61.9 & 75.7 & 77.5*\\
         \rowcolor{grayhighlight} \cellcolor{white}
         & \textbf{Ours} & {\bf 98.1} & 72.5 & 54.4 & {\bf 86.4} & {\bf 21.0} & {\bf 41.0} & {\bf 62.22} & {\bf 76.52} & {\bf 78.30} \\
        \midrule

        \multirow{4}{*}{Show-o-512~\cite{xieshow}} 
         & w/o SFT & 97.2 & 80.3 & 61.9 & 78.2 & 27.3 & 52.3 & 66.2 & 82.21 & 80.53\\
         & SFT*     & 97.2 & 84.1 & {\bf 66.6} & 80.9 & 26.8 & 46.8 & 67.03 & 81.8 & 80.02\\
         & Reca    & {\bf 98.1} & {\bf 93.4} & 64.7& 79.8 & {\bf 38.0} & {\bf 55.8} & {\bf 71.63}* & 84.94 & {\bf 84.47}\\
         
         \rowcolor{grayhighlight} \cellcolor{white}
         & \textbf{Ours} & 97.8 & 91.4 &  65.0 & {\bf 81.9} & 35.0 & 53.0 & 70.69 & {\bf 85.18} & 84.35\\
        \midrule

        \multirow{4}{*}{OpenUni-1.6B~\cite{wu2025openuni}} 
         & w/o SFT & 96.8 & 63.3 & 46.4 & 80.1 & 18.5 & 30.8 & 56.0 & 76.29 & 75.47 \\
         & SFT*    & 97.8 & 76.8 & 55.9 & 81.9 & 25.5 & 36.0 & 62.32 & 79.7  & 81.25\\
         & Reca    & 96.6 & {\bf 85.4} & 52.5& {\bf 84.3} & {\bf 46.5} & 50.8 & 69.33* &  80.45& 83.1*\\
         \rowcolor{grayhighlight} \cellcolor{white}
         & \textbf{Ours} & {\bf 98.8} & 84.6 & {\bf 56.9} & 84.0 & 38.8 & {\bf 54.0} &  {\bf 69.5} & {\bf 81.33} & {\bf 83.83} \\
        \midrule

        \multirow{4}{*}{OpenUni-3.6B~\cite{wu2025openuni}} 
         & w/o SFT & 99.1 & 71.8 & 51.9 & 83.9 & 23.3 & 41.6 & 61.9 & 79.02 & 78.84 \\
         & SFT*    & 98.8 & 81.3 & {\bf 55.9} & 85.9 & 25.0 & 48.8 & 65.94 & 80.45 & 82.6\\
         & Reca    & 99.1 & 92.7 & 52.3& {\bf 87.1} & 43.8 & 70.3 & 74.1 & 82.75 & 86.0*\\
         \rowcolor{grayhighlight} \cellcolor{white}
         & \textbf{Ours}  & {\bf 99.7} & {\bf 95.2} & 52.5 & 85.1 & {\bf 46.8} & {\bf 73.0} & {\bf 75.37} & {\bf 83.37}& {\bf 86.14} \\
        \midrule
        
        \multirow{4}{*}{Harmon-0.5B~\cite{wu2025harmonizing}} 
         & w/o SFT & 99.7 & 80.5 & 55.8 & 86.7 & 32.2 & 49.7 & 67.6 & 80.12 & 80.34\\
         & SFT*     & 100 & 86.4 & 64.4 & 87.5 & 37.8 & 56.5 & 72.1 & 82.5 & 83.5\\
         & Reca    & 99.9 & {\bf 92.3} &59.4 & {\bf 91.7} & 58.5 & 70.7 & 78.7 & 84.67 & 85.7*\\
         \rowcolor{grayhighlight} \cellcolor{white}
         & \textbf{Ours} & {\bf 100} & 90.2 & {\bf 65.3} & 91.5 & {\bf 66.0} & {\bf 75.8} & {\bf 81.5}  & {\bf 85.4}& {\bf 86.9} \\
        \midrule

        \multicolumn{11}{l}{\cellcolor{gray!10}\textbf{\textit{Fine-Tuning Dataset: BLIP3o-60k Dataset}}} \\ 
        \midrule
         & w/o SFT & 99.4 & 87.3 & 68.7 & 86.4 & 44.9 & 51.1 & 72.9 & 80.93 & 81.36\\
         & SFT*    & 99.1 & 94.7 & 77.8 & 86.4 & 78.3 & 74.0 & 85.0 & 85.6 & 87.32\\
         & Reca    & -- & -- & --&  --&  --& -- & 85.2 & 86.5 & 87.2* \\
         \rowcolor{grayhighlight} \cellcolor{white}
         \multirow{-4}{*}{Harmon-1.5B~\cite{wu2025harmonizing}} & \textbf{Ours} & {\bf 100} & {\bf 97.7} & {\bf 79.1} & {\bf 90.4} & {\bf 83.5} & {\bf 81.3} & {\bf 88.66}  & {\bf 86.58} & {\bf 88.08}\\
        \bottomrule
    \end{tabular}
    }
    \label{exp:main_experiments}
\end{table*}

\noindent{{\bf Datasets.}}
For fine-tuning, we combined high-quality generation data with an auxiliary instruction dataset, aligning with the experimental setup of Reca~\cite{xie2025reconstruction} for a fair comparison.
We used MidjourneyV6~\cite{midjourneyv6} and BLIP3o-60k~\cite{chen2025blip3} to improve aesthetic quality and semantic alignment, while incorporating LLaVA-Instruct-150K~\cite{liu2023visual} to preserve the model's understanding capability.

\setlength{\intextsep}{2pt} 
\setlength{\columnsep}{10pt}
\begin{wraptable}[7]{r}{0.55\textwidth}
    \centering
    \caption{Visual understanding performance.}
    \vspace{5pt}
    \setlength{\tabcolsep}{2pt}
    \renewcommand\arraystretch{0.95}
    \resizebox{\linewidth}{!}{
        \begin{tabular}{lcccccc}
            \toprule
            Method & MME & POPE$_{(\texttt{Acc})}$ & POPE$_{(\texttt{F1})}$ & GQA & MMMU & SEED \\
            \midrule
            w/o SFT & 1195 & 83.8 & 83.9 & \textbf{58.8} & 34.7 & 65.2 \\
            SFT & 1210 & 84.0 & 84.1 & 58.5 & 35.3 & 65.1 \\
            Reca & 1190 & 83.9 & 83.4 & 58.5 & 34.9 & 65.1 \\
            \rowcolor{grayhighlight} \cellcolor{white}
            Ours & {\bf 1217} & {\bf 84.5} & {\bf 84.2} & 58.7 & {\bf 36.0} & {\bf 65.5} \\
            \bottomrule
        \end{tabular}
    } 
    \label{tab.understanding}
\end{wraptable}
\noindent{{\bf Hyperparameter.}}
We set the softmax temperature $\tau=1.0$ for all softmax operations, text preservation ratio $\rho = 0.4$ in \eqref{eq:topk_text}, and joint objective weight $\lambda = 1.0$ in \eqref{eq:final_objective}. 
The visual hint selection ratio is $\eta = 0.3$ (or 0.4 for Show-o $512\times512$). 
Following prior UMMs~\cite{xieshow,wu2025harmonizing}, 
we adopt their masking schedules for sampling $\gamma(t)$.
\begin{figure*}[t]
    \centering
    \includegraphics[width=\linewidth]{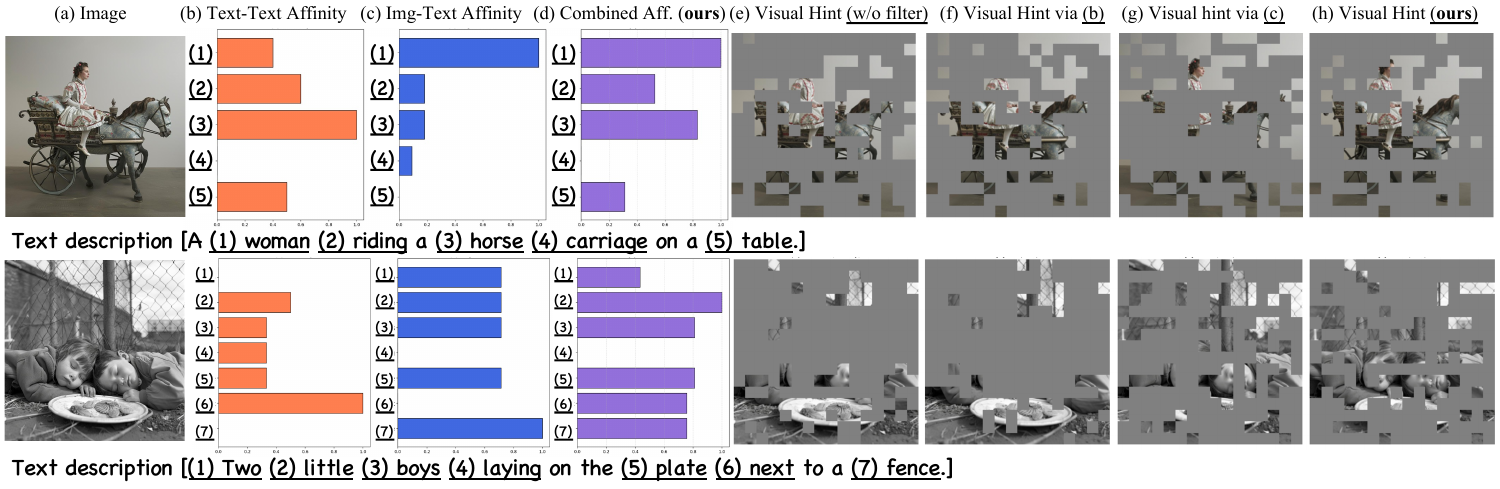}
    \caption{ 
         \textbf{Visualization of text token affinity and derived visual hints.} (a) The original input image with text description.
        (e--h) The extracted visual hints are represented by the image prompt. For (f--h), we project the text scores from (b--d) to compute the visual grounding map, retaining only the top 30\% of visual tokens with the highest groundedness scores. (e) shows the image prompt generated using all text words without discriminative filtering. Text scores exclude non-semantic functional stopwords.
    }
    \label{fig:qualitative_vis}
\end{figure*}

\subsection{Effectiveness of SeGroS in Text-to-Image Generation}
\begin{figure*}[t]
    \centering
    \includegraphics[width=\linewidth]{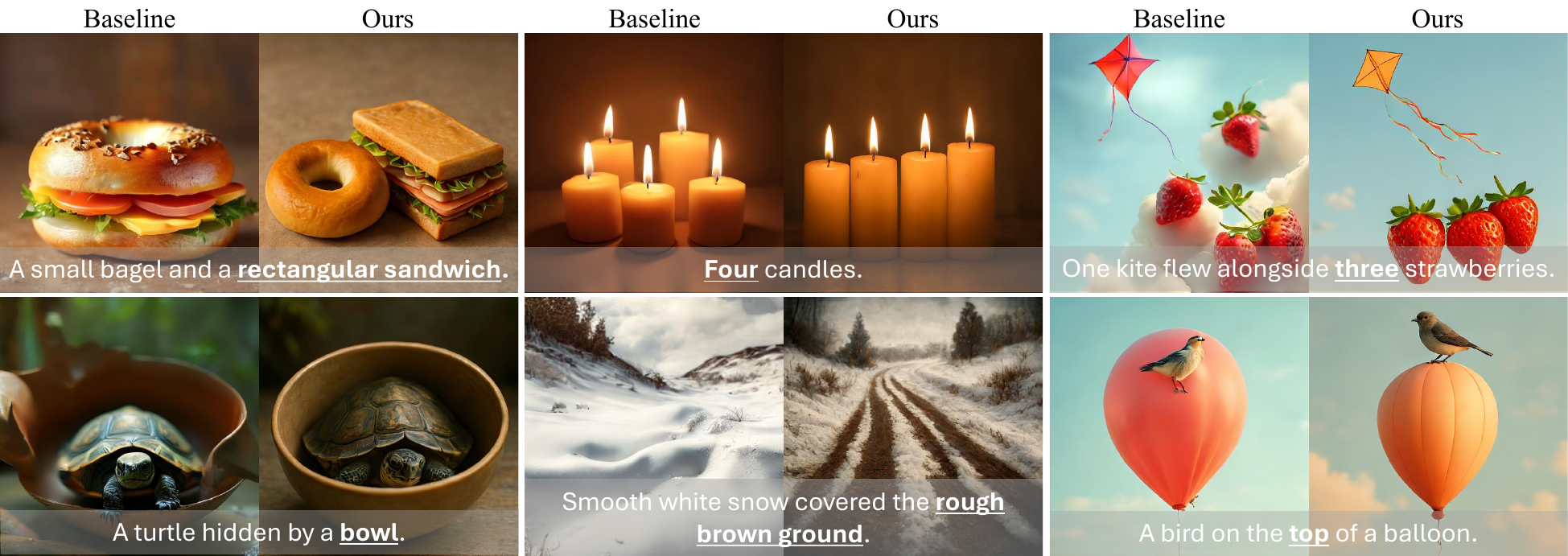}
    \caption{
         \textbf {Qualitative comparison of T2I results} using Harmon-1.5B~\cite{wu2025harmonizing} fine-tuned on the BLIP3o-60k~\cite{chen2025blip3} dataset. Baseline indicates the model before SFT. More visualizations for other backbones are provided in the supplementary material.
    }
    \label{fig:qualitative_result}
\end{figure*}

\begin{figure*}[t]
    \centering
    \includegraphics[width=\linewidth]{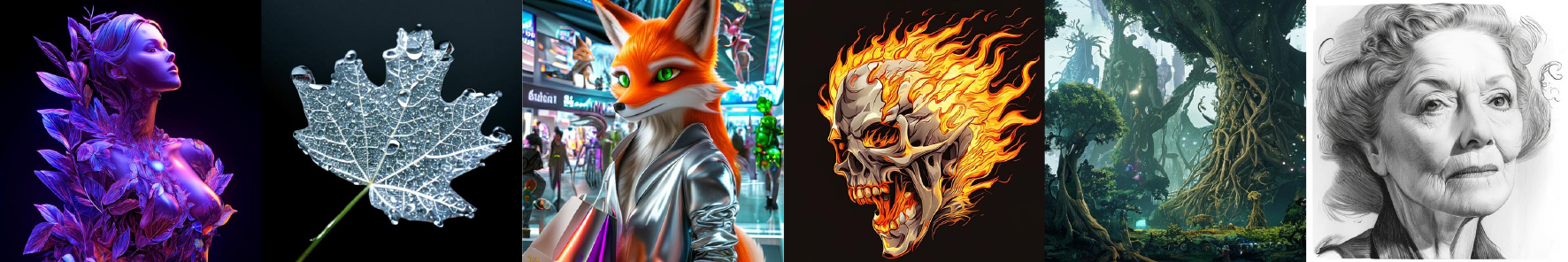}
    \caption{
         \textbf {Qualitative T2I generation results} of SeGroS with OpenUni-3.6B~\cite{wu2025openuni}. Refer to the supplementary material for the corresponding text prompts.
    }
    \label{fig:qualitative_result_openuni}
\end{figure*}

As shown in Tab.~\ref{tab:main_results}, SeGroS consistently improves upon standard supervised fine-tuning (SFT) across all evaluated UMM backbones and datasets, yielding higher overall scores on GenEval~\cite{ghosh2023geneval}, DPGBench~\cite{hu2024ella}, and CompBench~\cite{huang2023t2i} in most cases. Compared to Reca, SeGroS achieves higher DPGBench scores universally and superior GenEval overall scores in five of six settings (excepting a marginal drop on Show-o-512, where our DPGBench still leads). Notably, the gains are most pronounced in composition-sensitive GenEval categories (e.g., \textit{Position} and \textit{Attr.}), suggesting that our grounding-aware supervision effectively allocates learning capacity toward prompt-relevant structures. For instance, on OpenUni-3.6B, SeGroS improves the GenEval overall score from 65.94\% (SFT) and 74.1\% (Reca) to 75.37\%, while substantially boosting \textit{Position} accuracy (from 25.0\% to 46.8\%). We observe a similar trend when fine-tuning on BLIP3o-60k, where SeGroS pushes Harmon-1.5B to an 88.66\% GenEval overall score, indicating robustness across both backbone designs and data distributions. These quantitative trends are further supported qualitatively in Fig.~\ref{fig:qualitative_result}, where SeGroS better satisfies compositional constraints such as object counting and spatial relations. Fig.~\ref{fig:qualitative_result_openuni} further showcases its high-fidelity generation.

\subsection{Image-to-Text Understanding Results}
We evaluate image-to-text (I2T) understanding on the Harmon-1.5B~\cite{wu2025harmonizing} fine-tuned with the BLIP3o-60k dataset (Tab.~\ref{tab.understanding}).
Building upon Harmon's observation that improved generative modeling benefits visual understanding, we find that SeGroS effectively leverages this synergy.
Relative to w/o SFT, SeGroS yields a +22 gain on MME~\cite{fumme} and a +1.3 gain on MMMU~\cite{yue2024mmmu}, with consistent improvements on POPE~\cite{li2023evaluating} and SEED~\cite{li2023seed}. GQA~\cite{hudson2019gqa} performance remains stable (58.7 vs.\ 58.8), verifying the preservation of visual understanding capabilities.
Moreover, SeGroS consistently outperforms both standard SFT and the reconstruction-based Reca~\cite{xie2025reconstruction} variant across these benchmarks, indicating that grounding-aware supervision strengthens semantic representations.


\begin{table*}[t]
    \centering
    
    \begin{minipage}[t]{0.25\linewidth}
        \centering
        \caption{Effect of Visual Hint Ratio.}
        \label{tab:ablation_ratio_simple}
        \resizebox{1\linewidth}{!}{ %
            \begin{tabular}{c|c|c}
                \toprule
                \textbf{Ratio} ($\eta$) & \textbf{GenEval} & \textbf{DPGBench} \\
                \midrule
                $0\%$ & 72.1 & 82.5 \\
                $10\%$ & 80.4 & 84.5 \\
                $30\%$ & {\bf 81.5} & {\bf 85.4} \\
                $50\%$ & 81.2 & 85.1 \\
                $80\%$ & 79.8 & 84.9 \\
                $100\%$ & 79.2 & 84.5 \\
                \bottomrule
            \end{tabular}
        }
    \end{minipage}\hfill %
    \begin{minipage}[t]{0.35\linewidth}
        \centering
        \caption{Ablation on text filtering affinity.}
        \label{tab:ablation_text_affinity_components}
        \resizebox{0.75\linewidth}{!}{ 
            \begin{tabular}{c c | c c}
                \toprule
                \multicolumn{2}{c|}{\textbf{Metric}} & \multicolumn{2}{c}{\textbf{Performance}} \\
                \cmidrule(lr){1-2} \cmidrule(lr){3-4}
                \textbf{$\tilde{s}^{\text{intra}}$} & \textbf{$\tilde{s}^{\text{inter}}$} & \textbf{GenEval} & \textbf{DPGBench} \\
                \midrule
                -- & -- & 79.2 & 84.0 \\
                \checkmark & -- & 79.8 & 84.8 \\
                -- & \checkmark & 81.5 & 84.8 \\
                \midrule
                \checkmark & \checkmark &{\bf 81.5} & \textbf{85.4} \\
                \bottomrule
            \end{tabular}
        }
    \end{minipage} \hfill
    \begin{minipage}[t]{0.35\linewidth}
        \centering
        \caption{Ablation on visual hint and corrupted input.}
        \label{tab:ablation_roles_final}
        \resizebox{\linewidth}{!}{ 
            \begin{tabular}{cc|cc|cc}
                \toprule
                \multicolumn{2}{c|}{\textbf{Visual Hint}} & \multicolumn{2}{c|}{\textbf{$\mathbf{Z}_I^{\text{seen}}$}} & \multicolumn{2}{c}{\textbf{Performance}} \\
                \cmidrule(lr){1-2} \cmidrule(lr){3-4} \cmidrule(lr){5-6}
                \textbf{Top} & \textbf{Bot.} & \textbf{Top} & \textbf{Bot.} & \textbf{GenEval} & \textbf{DPGBench} \\
                \midrule
                -- & \checkmark & \checkmark & -- & 79.5 & 84.6 \\
                -- & \checkmark & -- & \checkmark & 80.4 & 84.5 \\
                \checkmark & -- & \checkmark & --  & 79.9 & 84.8 \\
                \checkmark & -- & -- & \checkmark  & \textbf{81.5} & \textbf{85.4} \\
                \bottomrule
            \end{tabular}
        }
    \end{minipage}

\end{table*}

\begin{table}[t]
    \centering
    \caption{
    \textbf{Ablation on supervision allocation:} 
    We form the corrupted input with a masking ratio $\gamma(t)\sim\mathcal{U}[0.7,1.0)$, \ie the ratio of visible tokens is $1-\gamma(t)$.
    While SFT employs random masking, our method adopts adaptive masking guided by the visual grounding map.     
    In `Ours (drop-loss)', supervision for the reconstruction loss during training is restricted to the Top-30\% grounded targets.
    }
    \label{tab:ablation_visible}
    \resizebox{0.9\linewidth}{!}{
    \begin{tabular}{c|c|c|c|c c}
        \toprule
        \textbf{Method} &
        \textbf{ Visual Hints } &
        \textbf{ Masking ratio } $\gamma(t)$ &
        \textbf{ Loss targets } &
        \textbf{ GenEval } $\uparrow$ &
        \textbf{ DPGBench } $\uparrow$ \\
        \midrule
        SFT & 0\%  & $\mathcal{U}[0.7,1.0)$ & All Masked Regions & 72.1 & 82.5 \\
        \midrule
        Ours & 30\% & $\mathcal{U}[0.7,1.0)$ & All Masked Regions & 81.5 & {\bf 85.4} \\
        \midrule
        Ours (drop-loss) & 30\% & $\mathcal{U}[0.7,1.0)$ & Top-$30\%$ Masked Regions & {\bf 82.1} & 84.7 \\
        \bottomrule
    \end{tabular}}
\end{table}

\subsection{Ablation study}
In this section, we conduct ablation studies to verify the effectiveness of each component in SeGroS. 
All ablation models are trained on the MidjourneyV6 dataset~\cite{midjourneyv6} using the Harmon-0.5B~\cite{wu2025harmonizing} backbone. 
We maintain the same hyperparameter settings as described in Tab.~\ref{exp:main_experiments}.

\subsubsection{Redundancy Analysis on Visual Hints.} We hypothesize that utilizing the entire image as a visual hint introduces spatial redundancy, diluting essential supervisory signals. To validate this, we ablate the visual hint ratio $\eta$ in Tab.~\ref{tab:ablation_ratio_simple}, where $\eta$ represents the percentage of top-scoring patches selected via Eq.~(\ref{eq:noise_score}). The results demonstrate that simply providing more visual information is not always beneficial; as $\eta$ increases from our optimal setting of $30\%$ to $100\%$ (full image), the performance steadily drops from 81.5 to 79.2 on GenEval. This confirms that uninformative patches introduce redundancy. Meanwhile, providing no hint ($\eta=0\%$) or an overly sparse hint ($\eta=10\%$) fails to provide sufficient structural guidance. Thus, retaining $30\%$ to $50\%$ of the highly-grounded patches strikes the optimal balance, preserving core semantics without capacity waste.

\subsubsection{Effectiveness of Text Filtering Components.}
We evaluate the individual and synergistic effects of intra-modal ($\tilde{s}^{\text{intra}}$) and inter-modal ($\tilde{s}^{\text{inter}}$) affinities (Tab.~\ref{tab:ablation_text_affinity_components}). The first row denotes the no-filtering image–text affinity baseline. Employing only the intra-modal score ($\tilde{s}^{\text{intra}}$) provides a structural linguistic prior that captures core semantic entities, steadily improving alignment (79.8\% / 84.8\%). Conversely, utilizing only the inter-modal score ($\tilde{s}^{\text{inter}}$) grounds the text tokens to visual patches, significantly boosting GenEval to 81.5\%. Finally, the best overall performance is achieved when both are combined (81.5\% / 85.4\%). 
Qualitatively, $\tilde{s}^{\text{intra}}$ tends to emphasize tightly coupled phrase-level tokens (e.g., \emph{riding--horse} in Fig.~\ref{fig:qualitative_vis} (b)), whereas $\tilde{s}^{\text{inter}}$ aggregates attention mass over patches and thus favors visually dominant entities (e.g., \emph{woman} in Fig.~\ref{fig:qualitative_vis} (c)); their combination yields the cleanest visual hints in Fig.~\ref{fig:qualitative_vis} (d,h).

\subsubsection{Ablation on Hint and Corrupted Input Construction.}
Tab.~\ref{tab:ablation_roles_final} ablates the assignment of high- (Top) and low-groundedness (Bot) patches to \textit{Visual Hints} and the unmasked context ($\mathbf{Z}_I^{\text{seen}}$).
Assigning Top patches to $\mathbf{Z}_I^{\text{seen}}$ misdirects the reconstruction loss toward irrelevant Bot regions, degrading performance (Rows 1, 3).
In contrast, retaining Bot patches as $\mathbf{Z}_I^{\text{seen}}$ correctly concentrates supervision on core semantics (Rows 2, 4).
Among these, providing Top rather than Bot patches as Visual Hints further enhances semantic conditioning (Row 4 vs.\ Row 2).
Overall, our default configuration (Row 4)—Top for Hints and Bot for $\mathbf{Z}_I^{\text{seen}}$—is optimal, successfully coupling informative visual guidance with semantically focused reconstruction.

\subsubsection{Validating the Supervisory Allocation.} 
We hypothesize that when constructing the corrupted input, the reconstruction loss should be strictly concentrated on semantically relevant regions, rather than being diluted across redundant areas.
To validate this, we fix the masking schedule $\gamma(t)\sim\mathcal{U}(0.7,1.0)$ for all settings and vary only \emph{where} the reconstruction loss is applied (Tab.~\ref{tab:ablation_visible}).
The SFT baseline randomly samples masked positions and supervises all masked tokens with the reconstruction loss.
In contrast, SeGroS uses the grounding map to select semantically aligned masked targets, yielding substantial gains in both GenEval and DPGBench.
We further test a drop-loss variant that restricts supervision to only the Top-$30\%$ grounded targets (assigning zero weight to the rest).
Despite backpropagating through only a small subset of masked positions, it remains strong—improving GenEval and still substantially outperforming SFT—while only mildly reducing DPGBench.
Overall, this suggests that grounding-aware allocation can retain most of the gains with far fewer supervised targets.

\section{Related work}
\noindent{{\bf Unified Multimodal Models (UMMs).}}
UMMs~\cite{liao2025mogao,liu2025tuna,wu2024vila} unify multimodal understanding and generation within a shared modeling framework. 
Early token-based approaches~\cite{team2024chameleon,chen2025janus} encode images as vector quantized tokens, enabling unified mixed-modal sequence modeling. 
To improve efficiency and representation quality, Show-o~\cite{xieshow} integrates discrete denoising diffusion with autoregressive modeling, while Harmon~\cite{wu2025harmonizing} adopts a MAR-based unified encoder for enhanced semantic consistency. OpenUni~\cite{wu2025openuni} bridges an autoregressive MLLM and a diffusion generator via a lightweight connector rather than fully sharing a single backbone. Some UMMs also leverage pretrained vision encoders (e.g., CLIP) as semantic priors for improved alignment~\cite{sun2023emu,wu2024vila,lin2025bifrost}.

\noindent{\bf Fine-tuning Strategies for UMMs.}
Fine-tuning plays a crucial role in improving UMM generation by refining text-to-visual alignment.
Reca~\cite{xie2025reconstruction} strengthens alignment through image-conditioned reconstruction, where the model reconstructs masked visual tokens from dense visual prompts.
Reward-driven alignment methods have also been explored, including self-rewarding frameworks and reinforcement-learning-based optimization~\cite{jin2025srum,mao2025unirl}.
BLIP3-o~\cite{chen2025blip3} further studies unified training recipes that incorporate supervised instruction tuning to improve alignment and aesthetics.
In contrast, SeGroS improves fine-tuning efficiency by grounding supervision in text--image correspondence and redesigning both conditioning cues and reconstruction targets to reduce prompt-irrelevant updates.

\noindent{{\bf Adaptive Masking Strategies.}}
Masked image modeling has become a prominent self-supervised paradigm for learning visual representations via masked reconstruction. 
However, random masking often allocates capacity to low-information background regions rather than object-level semantics~\cite{kakogeorgiou2022hide,li2022semmae,choi2024salience}. 
Prior work mitigates this limitation by masking salient or semantically coherent regions~\cite{kakogeorgiou2022hide,li2022semmae} or by adaptively adjusting masking ratios based on token salience~\cite{choi2024salience}. 
OneRef~\cite{xiao2024oneref} further explores text-guided masking in referring-centric pretraining settings to strengthen vision–language grounding.
While these methods improve representation learning through adaptive masking, they do not explicitly address how reconstruction supervision should be allocated in multimodal generation.

\section{Conclusion}
In this paper, we introduced SeGroS, a semantically-grounded fine-tuning framework that addresses the granularity mismatch and supervisory redundancy in UMMs. By leveraging a visual grounding map, SeGroS strategically formulates visual hints and corrupted input, concentrating the reconstruction loss exclusively on core text-aligned regions. As a result, SeGroS significantly improves cross-modal alignment across major benchmarks.

\bibliographystyle{splncs04}
\bibliography{main}
\newpage
\section{Overview}
We present additional details and results as follows:
\begin{itemize}
    \item Additional Implementation Details
    \item Additional Quantitative Results
    \item Additional Qualitative Results
	\item Backbone-Specific Objective Functions
	\item Discussion \& Limitations
    \item Text Prompts for Generated Images
\end{itemize}

\begin{table}[h]
    \centering
    \caption{\textbf{Hyperparameter settings for fine-tuning with SeGroS.} We provide the detailed configuration for reproducing our results across different UMM architectures.}
    \label{appendix:hyperparameter}
    \resizebox{0.8\linewidth}{!}{
        \begin{tabular}{lcccccc}
            \toprule
            & \multicolumn{2}{c}{\textbf{Show-o}~\cite{xieshow}} & \multicolumn{2}{c}{\textbf{Harmon}~\cite{wu2025harmonizing}} & \multicolumn{2}{c}{\textbf{OpenUni}~\cite{wu2025openuni}} \\
            \cmidrule(lr){2-3} \cmidrule(lr){4-5} \cmidrule(lr){6-7}
            \textbf{Configuration} & $\mathbf{256^2}$ & $\mathbf{512^2}$ & \textbf{0.5B} & \textbf{1.5B} & \textbf{1.6B} & \textbf{3.6B} \\
            \midrule
            \multicolumn{7}{l}{\textit{\textbf{Optimization}}} \\
            \quad Optimizer & AdamW & AdamW & AdamW & AdamW & AdamW & AdamW \\
            \quad Learning Rate & 5e-7 & 5e-7 & 1e-5 & 1e-5 & 1e-5 & 1e-5 \\
            \quad Weight Decay & 0.01 & 0.01 & 0.01 & 0.01 & 0.01 & 0.01 \\
            \quad Optimizer Momentum ($\beta_1, \beta_2$) & \multicolumn{2}{c}{(0.9, 0.999)} & \multicolumn{2}{c}{(0.9, 0.95)} & \multicolumn{2}{c}{(0.9, 0.95)} \\
            \midrule
            \multicolumn{7}{l}{\textit{\textbf{Training Schedule}}} \\
            \quad Training Steps & 5K & 5K & 3K & 5K & 5K & 5K \\
            \quad Warmup Steps & 1000 & 1000 & 500 & 500 & 50 & 50 \\
            \quad Batch Size & 4 & 2 & 96 & 48 & 42 & 20 \\
            \quad Gradient Accumulation & 5 & 20 & 2 & 2 & 16 & 16 \\
            \quad Training Time (h) & 4.5 & 20 & 3.5 & 7 & 1 & 2 \\
            \midrule
            \multicolumn{7}{l}{\textit{\textbf{SeGroS Specifics}}} \\
            \quad Visual Hint Ratio ($\eta$) & 30\% & 40\% & 30\% & 30\% & 30\% & 30\% \\
            \quad Text Filtering Ratio ($\rho$) & 40\% & 40\% & 40\% & 40\% & 40\% & 40\% \\
            \quad Loss Weight ($\lambda\mathcal{L}_{\text{i2t}}$) & 1.0 & 1.0 & 1.0 & 1.0 & 0.0 & 0.0 \\
            \quad Frozen Modules & \multicolumn{2}{c}{CLIP} & \multicolumn{2}{c}{-} & \multicolumn{2}{c}{MLLM} \\
            \bottomrule
        \end{tabular}
    }
\end{table}
\section{Experimental Setup}
\subsection{Training configuration}
All fine-tuning experiments across the various backbone models (i.e., Show-o~\cite{xieshow}, Harmon~\cite{wu2025harmonizing}, and OpenUni~\cite{wu2025openuni}) were conducted using a single NVIDIA H100 NVL (94GB) GPU. We explicitly utilize the native vision encoders and frozen internal text embedding layers inherent to each backbone, \textit{without introducing any external modules}.
The detailed training configurations are summarized in Tab.~\ref{appendix:hyperparameter}. This table includes the optimization settings, learning rate schedules, batch sizes, and SeGroS-specific hyperparameters applied to each model architecture. 
To ensure a strictly fair comparison, all baseline training hyperparameters are kept identical to those of Reca~\cite{xie2025reconstruction}.

\begin{figure*}[t] 
\begin{center}
\includegraphics[width=0.4\linewidth]{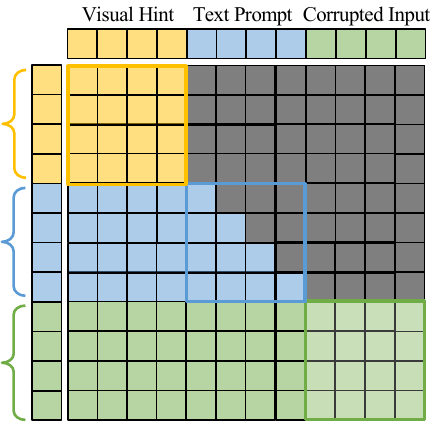}
\end{center}
\caption{\textbf{Attention mask adopted in SeGroS.} Dark cells denote blocked attention, whereas colored cells denote allowed attention. Rows correspond to query tokens and columns to key tokens.
}
\label{fig:attention_mask}
\end{figure*}
\subsection{Details of Input Sequence Construction}
We describe how Visual Hints are seamlessly integrated into the unified token sequence without requiring any backbone-specific architectural modifications.
As illustrated in Fig.~\ref{fig:attention_mask}, the Visual Hints are prepended to the sequence, followed immediately by the text prompt and the corrupted visual input. 
The Visual Hint tokens (yellow) attend exclusively within their own sequence in a bidirectional manner, while remaining independent of subsequent text or corrupted inputs.
The text tokens (blue) fully attend to the preceding Visual Hints while applying a standard causal mask over the text sequence.
Finally, the corrupted input tokens (green) have full bidirectional attention over the entire sequence, including the preceding Visual Hints and the text prompt.
Through this attention mechanism, the model can effectively leverage both the visual and textual guidance to process the corrupted image.\\

\subsection{Evaluation Setup and Benchmarks}
For the text-to-image evaluation phase, including both the generation of images and the computation of benchmark metrics, we utilized NVIDIA RTX A6000 (48GB) GPUs. We comprehensively evaluated our method across three benchmarks targeting compositional prompt adherence.

\noindent\textbf{GenEval~\cite{ghosh2023geneval}.}
We evaluate compositional prompt following on GenEval, which contains 553 templated prompts spanning six factors: single-object, two-object, counting, colors, spatial relations, and attribute binding. We report the aggregated verification score averaged across the 4 different random seeds to minimize variance.

\noindent\textbf{DPGBench~\cite{hu2024ella}.}
We assess long and dense prompt adherence on the Dense Prompt Graph Benchmark (DPG-Bench), which consists of 1,065 prompts with rich multi-entity attributes and relations. The reported overall DPG-Bench score is averaged across 4 different random seeds to ensure reliability.

\noindent\textbf{T2I-CompBench~\cite{huang2023t2i}.}
We further test compositional text-to-image generation on T2I-CompBench, a benchmark of 6,000 prompts covering attribute binding, object relations, spatial/non-spatial relations, and complex compositions. For this benchmark, the evaluation was conducted using images generated with a single seed. Following the recent evaluation protocol of Reca~\cite{xie2025reconstruction}, we employed the powerful Qwen2.5-VL-32B~\cite{bai2025qwen2} model as our visual-language model (VLM) evaluator to assess the compositional generation quality.

\section{Additional Quantitative Results}
\subsection{Effect of the Noise Scale in the Visual Grounding Map.}
In Tab.~\ref{tab:noise_ablation}, we investigate the effect of injecting uniform noise into the visual grounding scores before the Top/Bottom selection. Even without noise (Ours / None), SeGroS outperforms the SFT baseline, suggesting that grounding-guided selection is an important contributor. Introducing a moderate perturbation (noise scale $\alpha=0.5$) yields the best performance, demonstrating that incorporating randomness into the visual grounding map effectively improves the overall performance. Increasing the noise scale to $1.0$ slightly reduces the gain compared to $0.5$; however, it remains effective because the critical token scores are preserved to some extent despite the increased randomness.
\\

\begin{table}[t]
\centering
\caption{Ablation study on the noise scale $\alpha$ for the visual grounding map, where the noise vector is sampled as $\boldsymbol{\xi} \sim \mathcal{U}([0, \alpha]^{N_I})$ in Eq. (12).  Experiments are conducted using the Harmon-1.5B~\cite{wu2025harmonizing} backbone on BLIP3o-60k~\cite{chen2025blip3}.}
\label{tab:noise_ablation}
\resizebox{0.45\linewidth}{!}{\begin{tabular}{l|c |c c}
\toprule
\textbf{Method} & \textbf{Noise Scale ($\alpha$)} & \textbf{GenEval} & \textbf{DPGBench} \\
\midrule
w/o SFT & None & 72.9 & 80.9 \\
\midrule
SFT & None & 85.0 & 85.6 \\
\midrule
Ours & 0 (None) & 87.9 & 86.2 \\
Ours & 0.5 & \textbf{88.7} & \textbf{86.6} \\
Ours & 1.0 & 87.8 & 86.2 \\
\bottomrule
\end{tabular}}
\end{table}
\subsection{Transferability of the SeGroS Visual Grounding Map and Hint Redundancy}
As shown in Tab.~\ref{tab:comprehensive_ablation}, applying the visual grounding map to Reca yields consistent gains over vanilla Reca~\cite{xie2025reconstruction}: pruning dense visual hints from 100\% to the top 30\% increases GenEval (78.7 $\rightarrow$ 79.2) and DPGBench (84.67 $\rightarrow$ 85.2). This suggests that the visual grounding map serves as a transferable region-ranking prior that effectively reduces redundancy in dense visual hints and retains semantically salient patches. Building on this, the full SeGroS model achieves the best overall performance (81.5, 85.4), suggesting that the full gain arises from the joint effect of grounding-based hint selection, semantically grounded corrupted-input construction, and explicit text conditioning.

\begin{table}[t]
    \centering
    \caption{Ablation on visual hint redundancy and transferability of our visual grounding map. Applying the proposed visual grounding map to prune Reca's dense visual hints from 100\% to the top 30\% improves over vanilla Reca.}
    \label{tab:comprehensive_ablation}
    \resizebox{0.9\linewidth}{!}{
    \begin{tabular}{l|c|c|c|c|c|c}
        \toprule
        \textbf{Method} & \textbf{Text} & \textbf{Img.} & \textbf{Hint Selection Rule} & \textbf{Visual Hint Ratio} & \textbf{GenEval} & \textbf{DPG} \\
        \midrule
        Standard T2I & \checkmark & \texttimes & -- & 0\% & 72.08 & 82.45 \\
        \bottomrule
        Reca~\cite{xie2025reconstruction} & \texttimes & \checkmark & All visual hints  & 100\% & 78.7 & 84.67 \\
        Reca + SeGroS hint filtering & \texttimes & \checkmark & Top-30\% by SeGroS grounding map & 30\% & 79.2 & 85.2 \\
        \bottomrule
        \textbf{Ours (SeGroS)} & \checkmark & \checkmark & Top-30\% by SeGroS grounding map & 30\% & \textbf{81.5} & \textbf{85.4} \\
        \bottomrule
    \end{tabular}}
\end{table} 
\begin{table*}[t]
\centering
\caption{Detailed results on T2I-CompBench. Baseline denotes the original backbone models, and Ours denotes the proposed SeGroS framework. The experimental setup is identical to Tab. 1 in the main paper.}
\label{tab:compbench_breakdown}
\resizebox{0.65\linewidth}{!}{
\begin{tabular}{l | cc | cc | cc}
\toprule
\multirow{2}{*}{\textbf{Category}} & \multicolumn{2}{c|}{\textbf{Show-o-512}} & \multicolumn{2}{c|}{\textbf{Harmon-1.5B}} & \multicolumn{2}{c}{\textbf{OpenUni-3.6B}} \\  
& Baseline & Ours & Baseline & Ours & Baseline & Ours \\
\midrule
Color & 88.70 & \textbf{90.22} & 89.55 & \textbf{93.83} & 84.47 & \textbf{91.88} \\
Shape & 77.32 & \textbf{82.19} & 79.09 & \textbf{84.60} & 75.22 & \textbf{86.60} \\
Texture & 84.55 & \textbf{88.39} & 86.28 & \textbf{91.18} & 82.87 & \textbf{89.75} \\
Spatial & 83.96 & \textbf{87.05} & 84.98 & \textbf{93.19} & 78.29 & \textbf{86.95} \\
Non-spatial & 83.94 & \textbf{85.49} & 82.77 & \textbf{88.00} & 83.10 & \textbf{87.25} \\
Numeracy & 69.40 & \textbf{74.60} & 69.19 & \textbf{81.74} & 64.44 & \textbf{75.20} \\
3D Spatial & 74.90 & \textbf{82.85} & 75.58 & \textbf{86.77} & 68.60 & \textbf{85.18} \\
Complex & 81.47 & \textbf{83.97} & 82.05 & \textbf{85.31} & 82.82 & \textbf{86.29} \\
\midrule
\textbf{Overall} & 80.53 & \textbf{84.35} & 81.36 & \textbf{88.08} & 78.84 & \textbf{86.14} \\
\bottomrule
\end{tabular}
}
\end{table*}

\subsection{Detailed Quantitative Results}
Tab.~\ref{tab:compbench_breakdown} presents the detailed category-wise performance on T2I-CompBench~\cite{huang2023t2i}. Across all baseline models, SeGroS consistently improves generation quality in every single category without exception. Most notably, it yields significant gains in notoriously challenging areas such as Numeracy and 3D Spatial relationships, proving its robust compositional reasoning capabilities.

\section{Additional Qualitative Results}
\begin{figure*}[t]
    \centering
    \includegraphics[width=\linewidth]{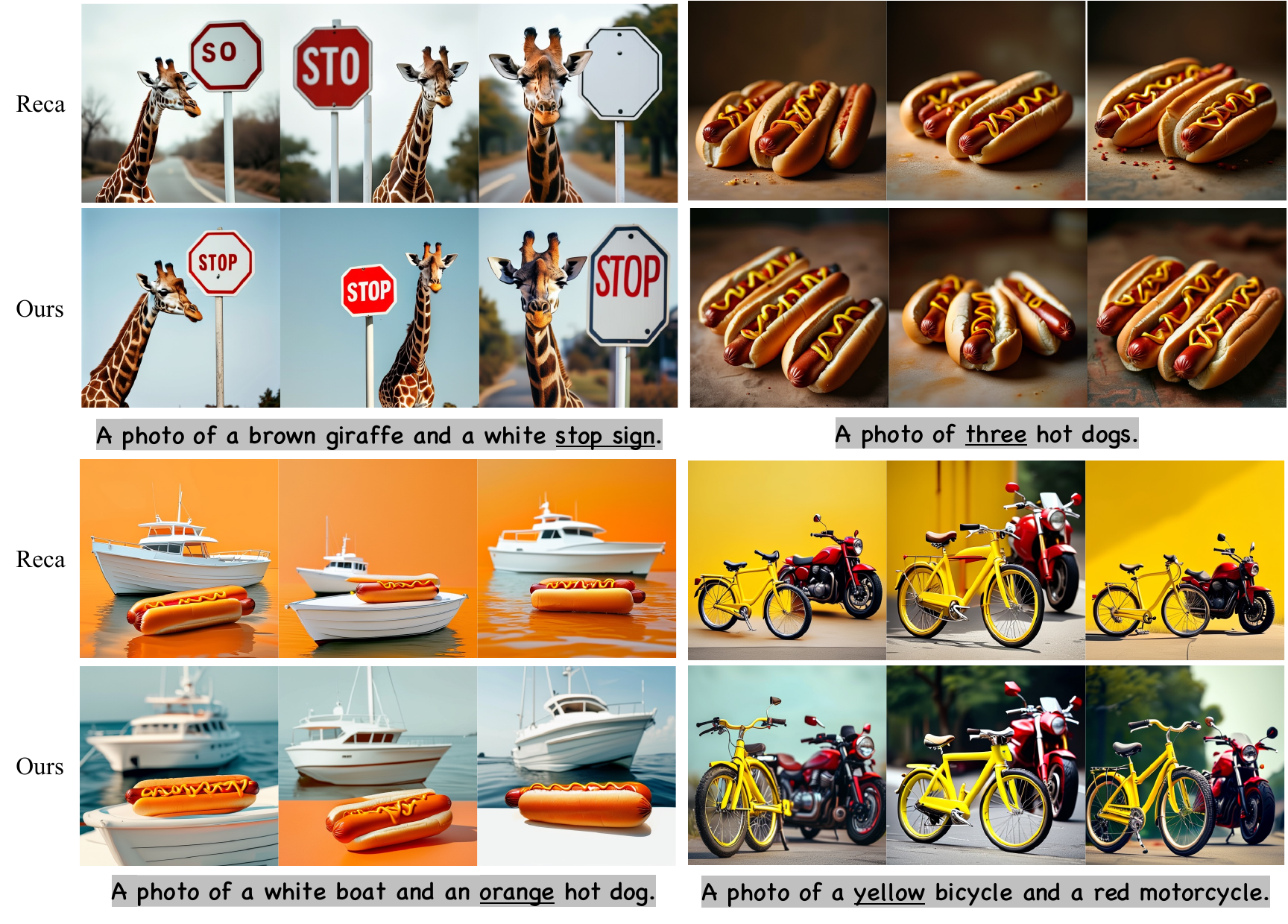}
    \caption{\textbf{Qualitative GenEval results on OpenUni-3.6B.}
We compare samples generated by Reca and SeGroS using the OpenUni-3.6B backbone under identical GenEval prompts.}
    \label{fig:openuni36_reca_ours}
\end{figure*}

\begin{figure*}[t]
    \centering
    \includegraphics[width=0.9\linewidth]{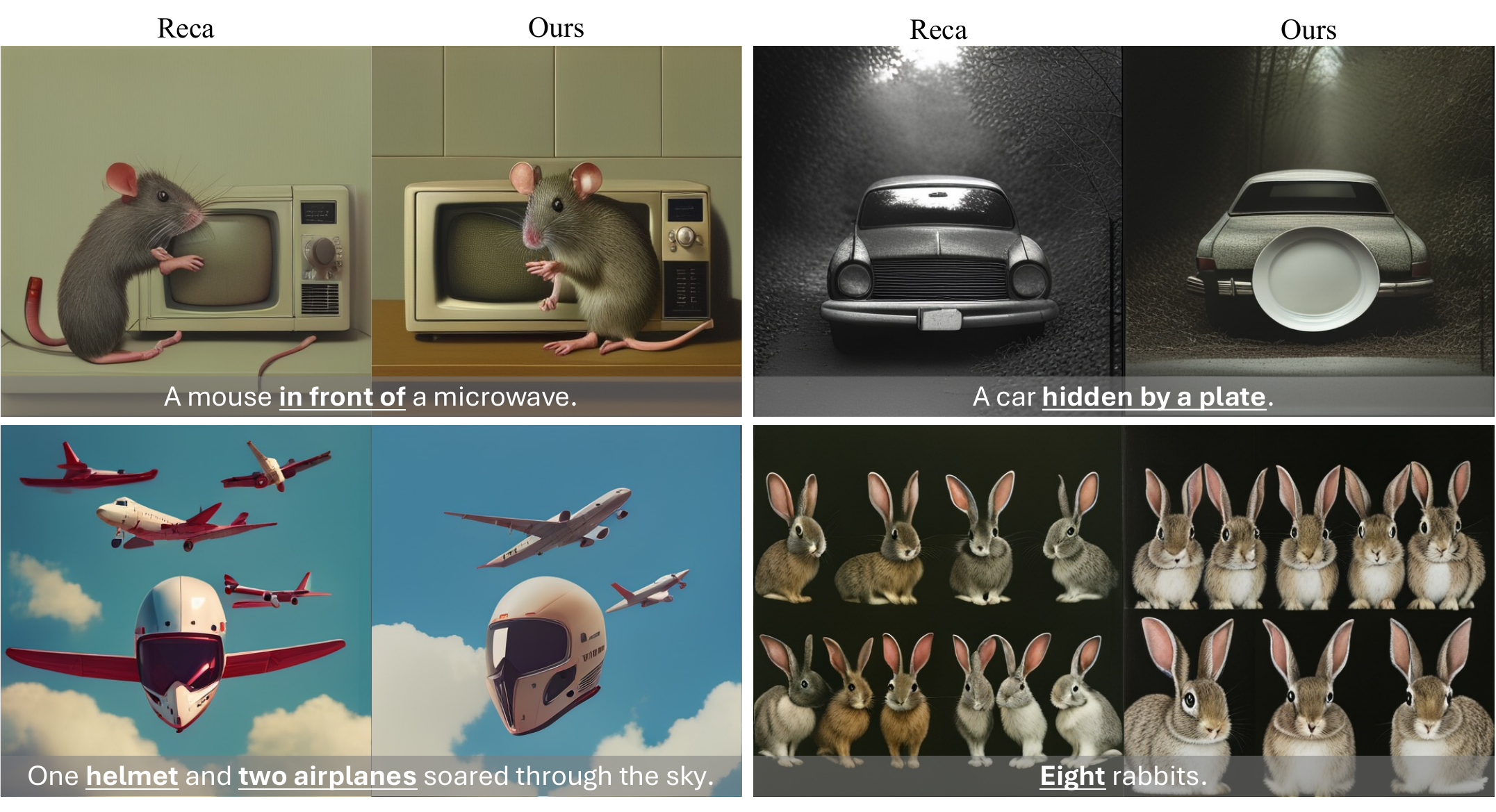}
    \caption{\textbf{Qualitative results on Show-o-512.}
We compare samples generated by Reca and SeGroS using the Show-o-512 backbone under identical T2I-CompBench prompts.}
    \label{fig:show512_reca_ours}
\end{figure*}

As seen in Fig.~\ref{fig:openuni36_reca_ours}, Reca often suffers from attribute binding failures (e.g., color bleeding) and incorrect object counting. In contrast, SeGroS ensures precise semantic alignment, maintaining structural integrity and prompt-faithful compositions even in complex multi-object scenarios. Furthermore, Fig.~\ref{fig:show512_reca_ours} demonstrates that this superiority extends to complex spatial relations and precise numeracy; unlike Reca, which struggles with occlusion and concept blending, SeGroS accurately interprets explicit spatial prepositions and strictly adheres to exact multi-object constraints.
\begin{figure*}[t]
    \centering
    \includegraphics[width=\linewidth]{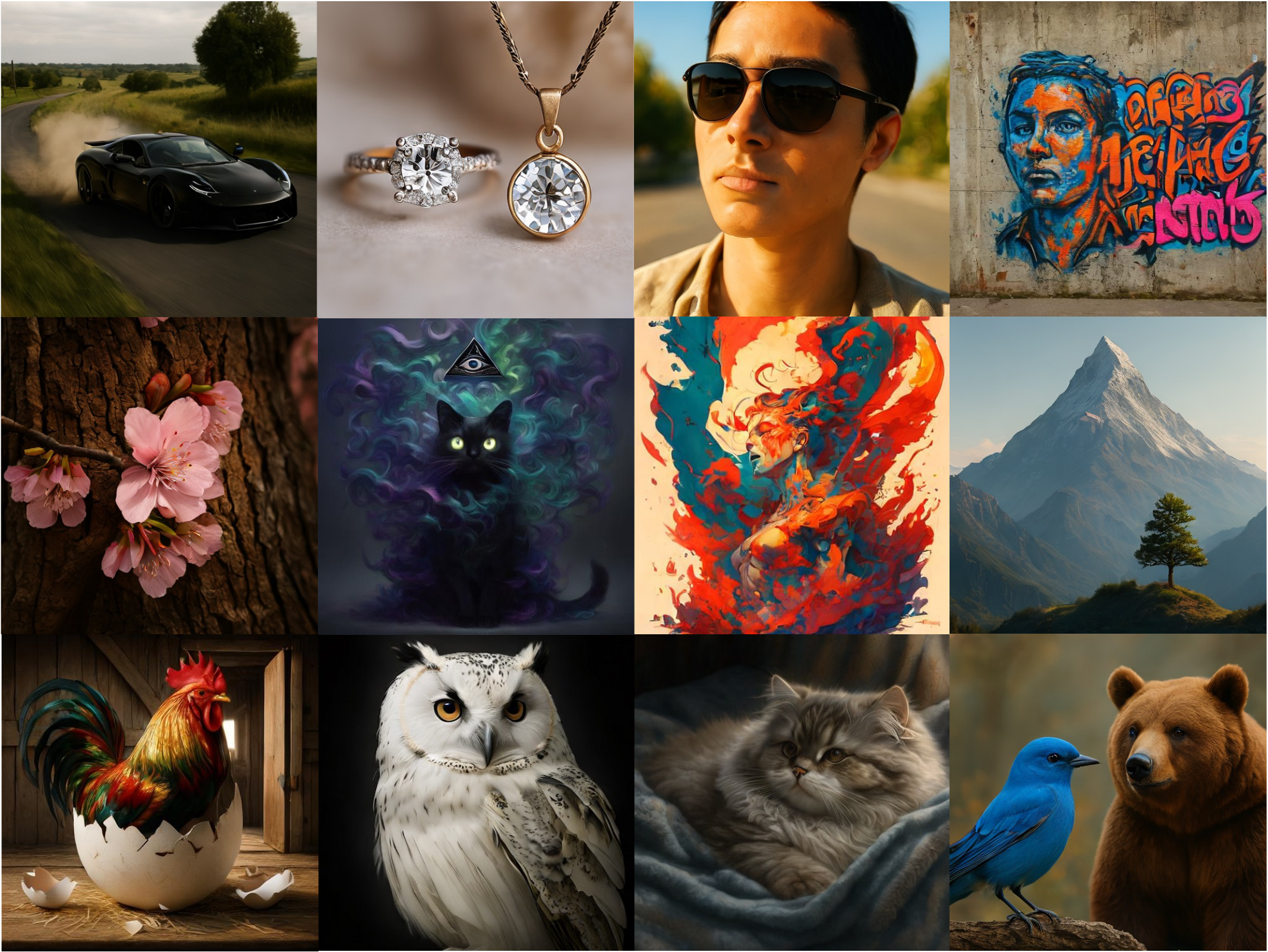}
    \caption{Qualitative results on Harmon 1.5B trained with SeGroS.}
    \label{fig:harmon15_ours}
\end{figure*}

\section{Backbone-Specific Objective Functions}
SeGroS is backbone-agnostic: it preserves each UMM backbone's native reconstruction objective and modifies only (i) the conditioning cues via text-grounded visual hints and (ii) the supervision allocation via semantically grounded masking. For a backbone $b$, we optimize
\begin{equation}
\mathcal{L}_{\mathrm{total}}^{(b)}
=
\mathcal{L}_{\mathrm{SeGroS}}^{(b)}
+
\lambda \mathcal{L}_{\mathrm{i2t}},
\end{equation}
where $\mathcal{L}_{\mathrm{SeGroS}}^{(b)}$ is evaluated only on the masked subset
$\mathcal{M}$ of the SeGroS corrupted input (with $|\mathcal{M}|$ following the
backbone's original masking schedule).

\subsubsection{Show-o backbone.}
Show-o~\cite{xieshow} represents an image as a sequence of discrete code indices
$\mathbf{v}_I=\{v_i\}_{i=1}^{N_I}$ with $v_i \in \{1,\dots,V\}$, which are mapped to token
embeddings for transformer processing.
SeGroS keeps Show-o's native masked token prediction objective and changes only the
construction of visual hints $\mathbf{Z}_I^{\mathrm{hint}}$ and the grounded corrupted
input $\widetilde{\mathbf{Z}}_I^{\mathrm{SeGroS}}$ (hence the masked set $\mathcal{M}$).
Formally, the SeGroS loss on Show-o is
\begin{equation}
\mathcal{L}_{\mathrm{SeGroS}}^{\mathrm{Show\mbox{-}o}}
=
-
\sum_{i \in \mathcal{M}}
\log p_{\theta}\!\left(
v_i \mid
\mathbf{Z}_T,\mathbf{Z}_I^{\mathrm{hint}},
\widetilde{\mathbf{Z}}_I^{\mathrm{SeGroS}}
\right),
\end{equation}
i.e., negative log-likelihood over the ground-truth code indices at masked positions.

\subsubsection{Harmon backbone.}
Harmon~\cite{wu2025harmonizing} performs masked generation in a MAR-style denoising framework over continuous
visual latents.
Let $\mathbf{z}^0\in\mathbb{R}^{N_I\times D}$ denote clean latent patches, and sample a
diffusion step $t$ and noise $\boldsymbol{\epsilon}\in\mathbb{R}^{N_I\times D}$.
Let $\mathbf{z}^t$ be the noised latents at step $t$, and let
$\widetilde{\mathbf{z}}^{t,\mathrm{SeGroS}}$ denote their semantically grounded corrupted
version (masking indices in $\mathcal{M}$ while keeping the rest as context) :
\begin{equation}
\widetilde{z}^{t,\mathrm{SeGroS}}_i =
\begin{cases}
z_i^t, & i\in\mathcal{M},\\
z_i^0, & i\notin\mathcal{M}.
\end{cases}
\end{equation}

With $\epsilon_{\theta}(\cdot)$ as Harmon's noise estimator conditioned on the text prompt
and visual hints, the SeGroS loss is

\begin{equation}
\mathcal{L}_{\mathrm{SeGroS}}^{\mathrm{Harmon}}
=
\mathbb{E}_{t,\boldsymbol{\epsilon}}
\left[
\sum_{i \in \mathcal{M}}
\left\|
\epsilon_i -
\epsilon_{\theta}\!\left(
\widetilde{\mathbf{z}}^{t,\mathrm{SeGroS}},\, t \mid
\mathbf{Z}_T,\mathbf{Z}_I^{\mathrm{hint}}
\right)_i
\right\|_2^2
\right],
\end{equation}
where $\epsilon_i$ and $(\cdot)_i$ denote the noise and prediction for the $i$-th latent
patch, respectively.
Thus, SeGroS does not replace Harmon's native denoising loss; it only reallocates which
patches are kept visible, used as hints, and reconstructed under the original objective.

\section{Discussion \& Limitations}
\noindent{\bf Discussion on the Show-o backbone.} 
SeGroS outperforms Reca on Show-o-256~\cite{xieshow}, validating our selective prompting strategy, while yielding modest gains on Show-o-512.
This attenuated improvement stems from the inherent architectural constraints of Show-o; specifically, its discrete tokenization~\cite{xie2025reconstruction} and the lack of high-level semantics in its CLIP encoder~\cite{tong2024eyes}.
We operate on the same backbone; it inevitably shares these constraints. 

\noindent{\bf Limitations.} 
SeGroS currently employs fixed ratios for text filtering and visual hint selection. While these fixed settings provide a robust baseline, dynamically tailoring the ratio per-instance could further unlock the framework's full potential. Moving towards adaptive selection strategies represents a compelling direction for future work.

\section{Text Prompts for Generated Images}
\subsection{Text Prompts for Fig.~\ref{fig:harmon15_ours}}
In Fig.~\ref{fig:harmon15_ours}, we present qualitative samples generated by our SeGroS-Harmon 1.5B~\cite{wu2025harmonizing} model. The corresponding text prompts for each image are listed below, ordered from left to right (and top to bottom):

\noindent{{\bf Image 1.}}
The sleek, black sports car sped down the winding country road, leaving a trail of dust in its wake.\\
\noindent{{\bf Image 2.}}
A diamond engagement ring and a round pendant.\\
\noindent{{\bf Image 3.}}
A person is wearing sunglasses on a sunny day.\\
\noindent{{\bf Image 4.}}
A textured concrete wall serves as a canvas for vibrant graffiti. The colorful artwork features a detailed portrait and bold lettering with various hues of blue, orange, and pink. Surrounding the graffiti, there are signs of age on the wall, including slight wear and chipped paint, highlighting the contrast between the old surface and the fresh paint.\\
\noindent{{\bf Image 5.}}
The soft pink petals of the cherry blossom contrasted with the rough brown bark.\\
\noindent{{\bf Image 6.}}
In the midst of an enigmatic scene, a mystical cat sits enveloped by undulating curls of vibrant smoke in hues of purple, blue, and green. Its eyes shine luminously, emitting an aura of enigmatic chaos magic that seems to dance around its sleek, shadowy fur. At the forefront of this visual marvel, an emblematic `all seeing eye' is prominently featured, adding to the arcane theme. The background fades into a blur, creating a shallow depth of field that places the focus squarely on the eerie yet captivating feline figure.\\
\noindent{{\bf Image 7.}}
This is a vibrant, digitally-created watercolor illustration portraying an apocalyptic scene with sharp focus and a smooth finish. The artwork, by James Jean, features Rossdraws' signature style with elements reminiscent of Frank Frazetta's fantasy aesthetics, incorporating Mcbess's bold linework, and infused with the ethereal quality of Sakimichan's enchantments. The dynamic composition showcases a whirlwind of colors that vividly depicts the chaotic yet mesmerizing moment at the end of the world.\\
\noindent{{\bf Image 8.}}
A big mountain and a small tree.\\
\noindent{{\bf Image 9.}}
A large, colorful rooster, with glossy feathers in shades of red, green, and gold, appears to be emerging from a cracked white eggshell. The scene unfolds on a rustic wooden table, with loose straw scattered around the egg's fragments. Against the backdrop, there's a barn door slightly ajar, allowing a sliver of daylight to accentuate the rooster's vibrant plumage.\\
\noindent{{\bf Image 10.}}
The soft white feathers of the owl contrasted with the sharp black talons.\\
\noindent{{\bf Image 11.}}
The fluffy cat was lying on the soft blanket.\\
\noindent{{\bf Image 12.}}
A blue bird and a brown bear.\\

\subsection{Text Prompts for Main Paper Fig. 5}
In Fig. 5 of the main manuscript, we presented qualitative text-to-image generation results using the OpenUni-3.6B~\cite{wu2025openuni} backbone fine-tuned with our SeGroS framework. These prompts are sourced from DPGBench~\cite{hu2024ella}, which uses dense, compositional descriptions to rigorously evaluate precise text-to-image alignment.

\noindent{{\bf Image 1.}} A visually striking digital portrayal of a futuristic woman, designed with an exquisite attention to detail that showcases her elegant pose, encased in an array of meticulously rendered leaves. Produced by the skilled artist Janice Sung, the image is a high-definition 4K creation that glows with stunning volumetric lighting effects. The woman's appearance is characterized by a level of hyper-realism that captures the intricate textures of her skin and the leaves, enhanced by the luminous, fantasy-inspired ambiance.

\noindent{{\bf Image 2.}} A close-up image capturing the intricate details of a maple leaf, which is composed entirely of clear, sparkling water droplets. The leaf is set against a smooth, dark background that accentuates its delicate water structure. The droplets glisten as they cling to the invisible veins of the leaf, creating a natural yet surreal piece of art.

\noindent{{\bf Image 3.}} A digital art piece depicting an anthropomorphic fox character with vibrant orange fur and emerald green eyes, wearing a sleek, silver-hued jacket and holding a neon-lit shopping bag, standing in the midst of a bustling, high-tech mall. The scene draws inspiration from Makoto Shinkai with its depth and use of light, the detailed and realistic texture akin to the works of James Gurney, while incorporating characteristic anime-style aesthetics. Surrounding the fox are other fantastical creatures, each rendered with the distinct, expressive touch reminiscent of Don Bluth's animation. Artists like Hibbary, Dark Natasha, and Goldenwolf could be credited with influencing the fox's lifelike fur and captivating poise. The image would be well-suited for the galleries of FurAffinity, known for its community of artists who celebrate anthropomorphic animals through their creative endeavors.

\noindent{{\bf Image 4.}} An intricately detailed representation of the Marvel character Ghost Rider featuring a human skull, with flames licking around the contours of the skull and rising above it in a fierce expression of fiery vengeance. The skull, alight with bright orange and yellow tones, dominates the image with its full head in view, set against a stark, void-like background that accentuates its fierceness. The character concept art captures both the macabre essence and supernatural intensity of the spectral figure.

\noindent{{\bf Image 5.}} A grand, sprawling landscape inspired by the iconic style of Hayao Miyazaki's Nausicaa of the Valley of the Wind and the Breath of the Wild from The Legend of Zelda series. The scene blends the fantastical elements of Studio Ghibli's post-apocalyptic setting with the vibrant, open-world aesthetic found in the game. Towering, ancient trees with twisted roots rise from the earth, while bioluminescent creatures add a touch of surreal luminance, hinting at an adventure awaiting at the edge of the world.

\noindent{{\bf Image 6.}}
An impressively detailed pencil illustration of Maggie Smith in the character of Reverend Mother is generating buzz on the ArtStation platform. The artwork, which has garnered awards for its lifelike quality, demonstrates a finesse reminiscent of Artgerm and Greg Rutkowski's dynamic strokes, with compositions that subtly hint at the influence of Alphonse Mucha's style. Its cinematic feel is accentuated by the careful play of light and shadow, earning acclaim and trending status among the art community.

\end{document}